\documentclass[conference]{IEEEtran}
\usepackage{times}

\usepackage[sorting=none, maxcitenames=1, maxbibnames=99, style=numeric-comp]{biblatex}
\usepackage{multicol}
\usepackage[bookmarks=true]{hyperref}
\usepackage{algorithm}
\usepackage[noend]{algpseudocode}
\usepackage{bm}
\usepackage{eqnarray}

\usepackage{graphicx}
\usepackage[font=small]{caption}
\usepackage{amsmath}
\usepackage{amssymb}
\usepackage{xcolor}
\hypersetup{colorlinks,linkcolor={black},citecolor={cyan},urlcolor={magenta}}

\usepackage{algorithm} 
\usepackage{algpseudocode} 
\usepackage{lipsum}
\usepackage{subcaption}

\usepackage{booktabs}
\newcommand{\ra}[1]{\renewcommand{\arraystretch}{#1}}

\usepackage{caption}
\begin{document}

\title{Sampling-based Exploration for\\ Reinforcement Learning of Dexterous Manipulation}




%
\author{\authorblockN{Gagan Khandate\authorrefmark{1}\authorrefmark{2},
Siqi Shang\authorrefmark{1}\authorrefmark{2},
Eric T. Chang\authorrefmark{3},
Tristan Luca Saidi\authorrefmark{2}, 
Yang Liu\authorrefmark{3}, \\
Seth Matthew Dennis\authorrefmark{3}, 
Johnson Adams\authorrefmark{3} and
Matei Ciocarlie\authorrefmark{3}}
\authorblockA{ \authorrefmark{2}Dept. of Computer Science~~~\authorrefmark{3}Dept. of Mechanical Engineering~~~\authorrefmark{1}joint first authorship\\
Columbia University, New York, NY 10027, USA
\\
Corresponding email:~\texttt{gagank@cs.columbia.edu}
}}

\maketitle

\begin{abstract}
In this paper, we present a novel method for achieving dexterous manipulation of complex objects, while simultaneously securing the object without the use of passive support surfaces. We posit that a key difficulty for training such policies in a Reinforcement Learning framework is the difficulty of exploring the problem state space, as the accessible regions of this space form a complex structure along manifolds of a high-dimensional space. To address this challenge, we use two versions of the non-holonomic Rapidly-Exploring Random Trees algorithm; one version is more general, but requires explicit use of the environment's transition function, while the second version uses manipulation-specific kinematic constraints to attain better sample efficiency. In both cases, we use states found via sampling-based exploration to generate reset distributions that enable training control policies under full dynamic constraints via model-free Reinforcement Learning. We show that these policies are effective at manipulation problems of higher difficulty than previously shown, and also transfer effectively to real robots. A number of example videos can also be found on the project website: \href{sbrl.cs.columbia.edu}{sbrl.cs.columbia.edu}

\end{abstract}

\IEEEpeerreviewmaketitle

\section{Introduction}

Reinforcement Learning (RL) of robot sensorimotor control policies has seen great advances in recent years, demonstrated for a wide range of motor tasks. In the case of manipulation, this has translated in higher levels of dexterity than previously possible, typically demonstrated by the ability to re-orient a grasped object in-hand using complex finger movements \cite{OpenAI2019-ng, Chen2021-ig, Qi2022-wy}. 

However, training a sensorimotor policy is still a difficult process, particularly for problems where the underlying state space exhibits complex structure, such as "narrow passages" between parts of the space are accessible or useful. Manipulation is indeed such a problem: even when starting with the object secured between the digits, a random action can easily lead to a drop, and thus to an irrecoverable state. Finger-gaiting further implies transitions between different subsets of fingers used to hold the object, all while maintaining stability. This leads to difficulty in exploration during training, since random perturbations in the policy action space are unlikely to discover narrow passages in state space. Current studies address this difficult through a variety of means: using simple, convex objects to reduce the difficulty of the task, reliance on support surfaces to reduce the chances of a drop, object pose tracking through extrinsic sensing, etc.

\begin{figure}[th!]
    \centering
    \includegraphics[width=\columnwidth]{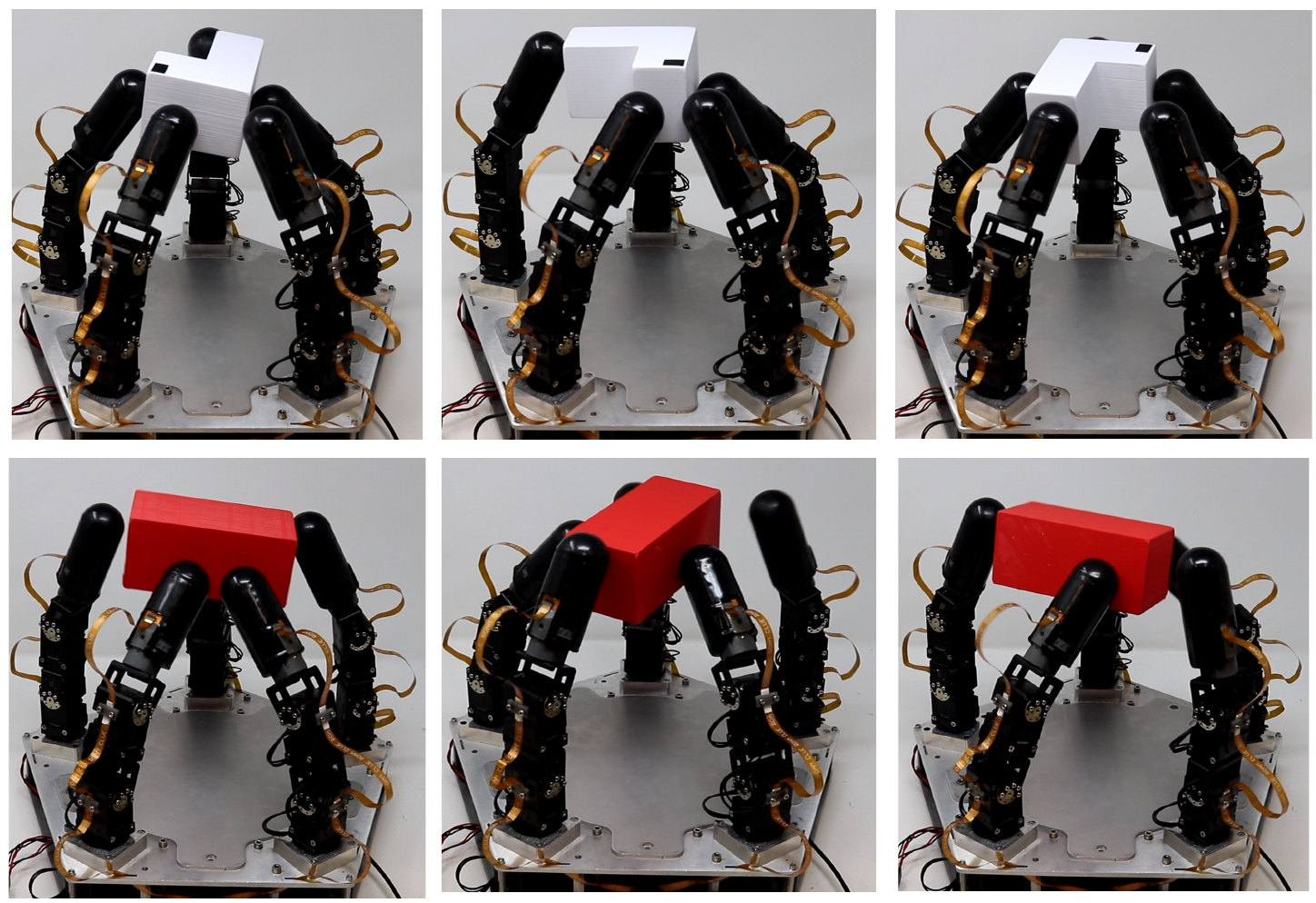}
    \caption{Our method enables finger-gaiting manipulation of concave or elongated objects which require complex gaits. We demonstrate these gaits both in simulation and on a real robot hand (shown here), using only proprioceptive and tactile feedback intrinsic to the hand.}
    \label{fig:eyecandy}
\end{figure}

The difficulty of exploring problems with labyrinthine state space structure is far from new in robotics. In fact, the large and highly effective family of Sampling-Based Planning (SBP) algorithms was developed in the field to address this exact problem. By expanding a known structure towards targets randomly sampled in the state space of the problem (as opposed to the action space of the agent), SBP methods can explore even very high-dimensional state spaces in ways that are probabilistically complete, or guaranteed to converge to optimal trajectories. However, SBP algorithms are traditionally designed to find trajectories rather than policies. For problems with computationally demanding dynamics, SBP can not be used on-line for previously unseen start states, or to quickly correct when unexpected perturbations are encountered along the way.

In this paper, we draw on the strength of both RL and SBP methods in order to train motor control policies for in-hand manipulation with finger gaiting. We aim to manipulate more difficult objects, including concave shapes, while securing them at all times without relying on support surfaces. Furthermore, we aim to achieve large re-orientation of the grasped object with purely intrinsic (tactile and proprioceptive) sensing. To achieve that, we explore multiple variants of the non-holonomic RRT algorithm with added constraints to find (approximate) trajectories that explore the useful parts of the problem state space. Then, we use these trajectories as reset distributions to train complete RL policies based on the full dynamics of the problem. Overall, the main contributions of this work include:
\begin{itemize}
    \item To the best of our knowledge, we are the first to show that reset distributions generated via SBP with kinematic constraints can enable more efficient training of RL control policies for dexterous in-hand manipulation.
    \item We show that SBP can explore useful parts of the manipulation state space by using either analytical approximations for contact and stability constraints, or by explicitly using the system's transition function (if available). When analytical approximations are used, RL later fills in the gaps by learning appropriate actions under more realistic dynamic constraints.
    \item The exploration boost from SBP allows us to train policies for dexterous skills not previously shown, such as in-hand manipulation of concave shapes, with only intrinsic sensing and no support surfaces. We demonstrate these skills both in simulation and on real hardware.
\end{itemize}

\section{Related Work}
 
Exploration methods for general RL operate under the strict assumption that the learning agent cannot teleport between states, mimicking the constraints of the real world. Under such constraints, proposed exploration methods include using intrinsic rewards \cite{Pathak2017-ik, Haarnoja2018-zj} or improving action consistency via temporally correlated noise in policy actions  
\cite{Amin2021-dm} or parameter space noise \cite{ Plappert2017-ba}.  

Fortunately, in cases where the policies are primarily trained in simulation, this requirement can be relaxed, and we can use our knowledge of the relevant state space to design effective exploration strategies. A number of these methods improve exploration by injecting useful states into the reset distribution during training. \textcite{Nair2017-nu} use states from human demonstrations in a block stacking task, while Ecoffet et al. \cite{Ecoffet2019-lk, Ecoffet2021-xs} use states previously visited by the learning agent itself for problems such as Atari games and robot motion planning. \textcite{Tavakoli2018-ah} evaluate various schemes for maintaining and resetting from the buffer of visited states. However, these schemes were evaluated only on benchmark continuous control tasks \cite{Duan2016-em}. From a theoretical perspective, \textcite{Agarwal2020-zu} show that a favorable reset state distribution provides a means to circumvent worst-case exploration issues, using sample complexity analysis of policy gradients. 

Finding feasible trajectories through a complex state space is a well-studied motion planning problem. Of particular interest to us are sampling-based methods such as Rapidly exploring Random Trees (RRT) \cite{LaValle1998-kn, Karaman2010-zi, Webb2013-nt} and Probabilistic Road Maps (PRM) \cite{Kavraki1996-gr, Kavraki1998-wl}. These families of methods have proven highly effective, and are still being expanded. Stable Sparse-RRT (SST) and its optimal variant SST* \cite{Li2021-lv} are examples of recent sampling-based methods for high-dimensional motion planning with physics. However, the goal of these methods is finding (kinodynamic) trajectories between known start and goal states, rather than closed-loop control policies which can handle deviations from the expected states.


Several approaches have tried to combine the exploratory ability of SBP with RL, leveraging planning for global exploration while learning a local control policy via RL \cite{Chiang2019-vm,Francis2020-qe,Schramm2022-hs}. These methods were primarily developed for and tested on navigation tasks, where nearby state space samples are generally easy to connect by an RL agent acting as a local planner. The LeaPER algorithm \cite{Pinto2018-xi} also uses plans obtained by RRT as reset state distribution and learns policies for simple non-prehensile manipulation. However, the state space for the prehensile in-hand manipulation tasks we show here is highly constrained, with small useful regions and non-holonomic transitions. Other approaches use trajectories planned by SBP as expert demonstrations for RL~\cite{Morere2020-gq}, but this requires that planned trajectories also include the actions used to achieve transitions, which SBP does not always provide. Alternatively, \textcite{Jurgenson2019-ye} and \textcite{Ha2020-od} use planned trajectories in the replay buffer of an off-policy RL agent for multi-arm motion planning. However, it is unclear how off-policy RL can be combined with the extensive physics parallelism that has been vital in the recent success of on-policy methods for learning manipulation \cite{Allshire2021-qp, Makoviychuk2021-ko, Chen2021-ig}. 



Turning specifically to the problem of dexterous manipulation, a number of methods have been used to advance the state of the art, including planning, learning, and leveraging mechanical properties of the manipulator. \textcite{Leveroni1996-iy} build a map of valid grasps and use search methods to generate gaits for planar reorientation, while \textcite{Han1998-xj} consider finger-gaiting of a sphere and identify the non-holonomic nature of the problem. Some methods have also considered RRT for finger-gaiting in-hand manipulation \cite{Yashima2003-lw, Xu2007-yb}, but limited to simulation for a spherical object. More recently, Morgan et al. demonstrate robust finger-gaiting for object reorientation using actor-critic reinforcement learning~\cite{Morgan2021-ny} and multi-modal motion planning~\cite{Morgan2022-xt}, both in conjunction with a compliant, highly underactuated hand designed explicitly for this task. \textcite{Bhatt2022-fr} also demonstrate robust finger-gaiting finger-pivoting manipulation with a soft compliant hand, but these skills were not autonomously learned but rather hand-designed and executed in an open-loop fashion.

Model-free RL has also led to significant progress in dexterous manipulation, starting with {OpenAI's} demonstration of finger-gaiting and finger-pivoting~\cite{OpenAI2019-ng}, trained in simulation and translated to real hardware. However, this approach uses extensive extrinsic sensing infeasible outside the lab, and relies on support surfaces such as the palm underneath the object. \textcite{Khandate2022-qt} show dexterous finger-gaiting and finger-pivoting skills using only precision fingertip grasps to enable both palm-up and palm-down operation, but only on a range of simple convex shapes and in a simulated environment. \textcite{Makoviychuk2021-ko} showed that GPU physics could be used to accelerate learning skills similar to OpenAI's. \textcite{Allshire2021-qp}  used extensive domain randomization and sim-to-real transfer to re-orient a cube but used table top as an external support surface. \textcite{Chen2021-ig, Chen2022-ud} demonstrated in-hand re-orientation for a wide range of objects under palm-up and palm-down orientations of the hand with extrinsic sensing providing dense object feedback. \textcite{Sievers2022, Pitz2023} demonstrated in-hand cube reorientation to desired pose with purely tactile feedback. \textcite{Qi2022-wy} used rapid motor adaptation to achieve effective sim-to-real transfer of in-hand manipulation skills for small cylindrical and cube-like objects. In our case, the exploration ability of SBP allows learning of policies for more difficult tasks, such as in-hand manipulation of non-convex and large shapes, with only intrinsic sensing. We also achieve successful, robust sim-to-real transfer without extensive domain randomization or domain adaptation, by closing the sim-to-real gap via tactile feedback.

\section{Method}

In this paper, we focus on the problem of achieving dexterous in-hand manipulation while simultaneously securing the manipulated object in a precision grasp. Keeping the object stable in the grasp during manipulation is needed in cases where a support surface is not available, or the skill must be performed under different directions for gravity (i.e. palm up or palm down). However, it also creates a difficult class of manipulation problems, combining movement of both the fingers and the object with a constant requirement of maintaining stability. In particular, we focus on the task of achieving large in-hand object rotation, which we, as others before~\cite{Qi2022-wy}, believe to be representative of this general class of problems, since it requires extensive finger gaiting and object re-orientation. 

\subsection{Problem Description}
\label{sec:nonhol}

Formally, our goal is to obtain a policy for issuing finger motor commands, rewarded by achieving large object rotation around a given hand-centric axis. The state of our system at time $t$ is denoted by $\bm{x}_t=(\bm{q}_t, \bm{p}_t)$, where $\bm{q} \in \mathcal{R}^d$ is a vector containing the positions of the hand's $d$ degrees of freedom (joints), and $\bm{p} \in \mathcal{R}^6$ contains the position and orientation of the object with respect to the hand. An action (or command) is denoted by the vector $\bm{a} \in \mathcal{R}^d$ comprising new setpoints for the position controllers running at every joint.

For parts of our approach, we assume that a model of the forward dynamics of our environment (\textit{i.e.} a physics simulator) is available for planning or training. We denote this model by $\bm{x}_{t+1} = F(\bm{x}_t, \bm{a}_t)$. We will show however that our results transfer to real robots using standard sim-to-real methods.  

We chose to focus on the case where the only sensing available is hand-centric, either tactile or proprioceptive. Achieving dexterity with only proprioceptive sensing, as biological organisms are clearly capable of, can lead to skills that are robust to occlusion and lighting and can operate in very constrained settings. With this directional goal in mind, the observation available to our policy consists of tactile and proprioceptive data collected by the hand, and no global object pose information. Formally,  the observation vector is
\begin{equation}
\bm{o}_t = [\bm{q}_t, \bm{q}^s_t, \bm{c}_t]    
\end{equation}
where $\bm{q}_t, \bm{q}^s_t \in \mathcal{R}^d$ are the current positions and setpoints of the joints, and $\bm{c}_t \in [0, 1]^m$ is the vector representing binary (contact / no-contact) touch feedback for each of $m$ fingers of the hand.

As discussed above, we also require that the hand maintain a stable precision grasp of the manipulated object at all times. Overall, this means that our problem is characterized by a high-dimensional state space, but only small parts of this state space are accessible for us: those where the hand is holding the object in a stable precision grasp. Furthermore, the transition function of our problem is non-holonomic: the subset of fingers that are tasked with holding the object at a specific moment, as well as the object itself, must move in concerted fashion. Conceptually, the hand-object system must evolve on the complex union of high-dimensional manifolds that form our accessible states. Still, the problem state space must be effectively explored if we are to achieve dexterous manipulation with large object re-orientation and finger gaiting.

\subsection{Manipulation RRT}
\label{sec:manRRT}

To effectively explore our high-dimensional state space characterized by non-holonomic transitions, we turn to the well-known Rapidly-Exploring Random Trees (RRT) algorithm. We leverage our knowledge of the manipulation domain to induce tree growth along the desired manifolds in state space. In particular, we expect two conditions to be met for any state: (1) the hand must maintain at least three fingers in contact with the object\footnote{Three contacts are the fewest that can achieve stable grasps without relying on torsional friction, which is highly sensitive to the material properties of the objects in contact. In the future, we plan to also include two-contact conditions, which can allow a richer set of manipulation primitives, at the expense of more complex contact modeling.}, and (2) the distribution of these contacts must be such that a stable grasp is possible. We note that these are necessary, but not sufficient conditions for stability; nevertheless, we found them sufficient for effective exploration.

Preservation of condition (1) during the transition between two states means that the object and the fingers that maintain contact with it must move in unison. Assume that we would like the system to evolve from state $\bm{x}_{start}=(\bm{q}_{start},\bm{p}_{start})$ towards state $\bm{x}_{end}(\bm{q}_{end},\bm{p}_{end})$, with a desired change in state of $\Delta \bm{x}_{des} = (\Delta \bm{q}_{des}, \Delta \bm{p}_{des}) = \bm{x}_{end} - \bm{x}_{start}$. Further assume that the set $S$ comprises the indices of the fingers that are expected to maintain contact throughout the motion. The requirement of maintaining contact, linearized around $\bm{x}_{start}$, can be expressed as:
\begin{equation}
    \bm{J}_S(\bm{q}_{start})\Delta \bm{q}_{des} = \bm{G}_S(\bm{p}_{start})\Delta \bm{p}_{des}
\end{equation}
where $\bm{J}_S(\bm{q}_{start})$ is the Jacobian of contacts on fingers in set $S$ computed at $\bm{q}_{start}$, and $\bm{G}_S(\bm{p}_{start})$ is the grasp map matrix of contacts on fingers in set $S$ computed at $\bm{p}_{start}$. This is further equivalent to
\begin{equation}
    \bm{N}_S(\bm{x}_{start})\Delta \bm{x}_{des} = 0
\end{equation}
where $\bm{N}_S(\bm{x}_{start})=\left [ \bm{J}_S(\bm{q}_{start})~~~-\bm{G}_S(\bm{p}_{start}) \right ]^T$. 

It follows that, if the desired direction of motion in state space $\Delta\bm{x}_{des}$ violates this constraint, we can still find a similar movement that does not violate the constraint by projecting the desired vector into the null space of the matrix $\bm{N}$ as defined above:
\begin{align}
    \Delta \bm{x}_{proj} &= (\bm{I}-\bm{N}^T\bm{N})\Delta\bm{x}_{des} \label{eq:project} \\
    \bm{x}_{new} &= \bm{x}_{start} + \alpha \Delta \bm{x}_{proj} \label{eq:newstate}
\end{align}
where $\alpha$ is a constant determining the size of the step we are willing to take in the projected direction.

We note that this simple projection linearizes the contact constraint around the starting state. Even for small $\alpha$, small errors due to this linearization can accumulate over multiple steps leading the fingers to lose contact. Thus, in practice, we further modify $\bm{x}_{new}$ by bringing back into contact with the object any finger that is within a given distance threshold (in practice, we set this threshold to 5 mm).

Maintaining at least three contacts with the object does not in itself guarantee a stable grasp. We take further steps to ensure that the contact distribution is appropriate for stability. Assume a set of $k$ contacts, where each contact $i$ has a normal direction $\bm{n}_i$ expressed in the global coordinate frame. We require that, if at least one contact $j$ applies a non-zero normal contact force of magnitude $c_j$, the other contacts must be able to approximately balance it via normal forces of their own, minimizing the resulting net wrench applied to the object. This is equivalent to requiring that the hand have the ability to create internal object forces by applying normal forces at the existing contacts. We formulate this problem as a Quadratic Program:
\begin{align}
\text{unknowns: normal}&\text{ force magnitudes }c_i,~i=1 \ldots k \nonumber \\
\text{minimize}~\|\bm{w}\|&~\text{subject to:} \nonumber \\
\bm{w} &= \bm{G}^T \left[ c_1 \bm{n_1} \ldots c_k \bm{n_k} \right]^T \label{eq:stabstart}\\
c_i & \geq  0~\forall i \\
\exists j~\text{such that}~c_j &= 1~\text{(ensure non-zero solution)}
\end{align}
If the resulting minimization objective is below a chosen stability threshold, we deem the grasp to be stable:
\begin{equation}
    \text{If }\|\bm{w}\| < \epsilon_{stab}\text{: grasp is stable} \label{eq:stabend}
\end{equation}
We note that this measure is conservative in that it does not rely on friction forces. Furthermore, it ensure that the fingers are able to generate internal object forces using contact normal forces, but does not specify what are appropriate motor torques for doing so. Nevertheless, we have found it effective in pushing exploration towards useful parts of the state space.

\begin{algorithm}[t]
\caption{Manipulation RRT (M-RRT)}\label{alg:manipulation}
\begin{algorithmic}[1]
\Require Tree contains root node; $N \gets 1$
\While{$N<N_{max}$}
\State $\bm{x}_{sample} \gets$ random point in state space
\State $\bm{x}_{node} \gets$ node closest to $\bm{x}_{sample}$ currently in tree
\State $\Delta \bm{x}_{des} \gets \|\bm{x}_{sample} - \bm{x}_{node}\|$
\State $\mathcal{S} \gets$ all sets of three fingers contacting the object in state $\bm{x}_{node}$ 
\State $d_{min} \gets \infty; x_{new} \gets $ NULL
\ForAll{$S_i$ in $\mathcal{S}$} \label{line:propstart}
\State Compute $\bm{x}_i$ by projecting $\Delta\bm{x}_{des}$ on the constraint manifold of contacts in $S_i$ as in eqs.~(\ref{eq:project}-\ref{eq:newstate}) \label{line:project}
\If{Stable($x_i$) \textbf{and} $dist(x_{sample},x_i) < d_{min}$ }\label{line:stable2}
\State $d_{min} \gets dist(x_{sample},x_i)$
\State $x_{new} \gets x_i$ \label{line:propend}
\EndIf
\EndFor
\If{$x_{new}$ is not NULL}
\State Add $x_{new}$ to tree with $x_{node}$ as parent
\State $N\gets N+1$
\EndIf
\EndWhile
\end{algorithmic}
\end{algorithm}

We can now put together these constraints into the complete algorithm shown in Alg.~\ref{alg:manipulation} and referred to in the rest of this paper as M-RRT. The essence of this algorithm is the forward propagation in lines \ref{line:propstart}-\ref{line:propend}. Given a desired direction of movement in state space, we want to ensure that at least three fingers maintain contact with the object. We thus project the direction of motion onto each of the manifolds defined by the contact constraints of each possible set of three fingers that begin the transition in contact with the object. We then choose the projected motion that brings us closest to the desired state-space sample. Finally, we perform an analytical stability check on the new state in line \ref{line:stable2} via eqs.~(\ref{eq:stabstart}-\ref{eq:stabend}).

We note that M-RRT does not make use of the environment's transition function $F()$ (i.e. system dynamics). In fact, both the projection method in eqs.~(\ref{eq:project}-\ref{eq:newstate}) and the stability check via eqs.~(\ref{eq:stabstart}-\ref{eq:stabend}) can be considered as approximations of the transition function, aiming to preserve movement constraints but without explicitly computing and checking the system's dynamics. As such, they are fast to compute but approximate in nature. It is possible that some of the transitions in the resulting RRT tree are in fact invalid under full system dynamics, or require complex sequences of motor actions. As we will see in Sec.~\ref{sec:RL} however, they are sufficient for helping learn a closed-loop control policy. Furthermore, for cases where the $F()$ is available and fast to evaluate, we also study a variant of our approach that makes explicit use of it in the next section.

\subsection{General-purpose non-holonomic RRT}

For problems where system dynamics $F()$ are available and fast to evaluate, we also investigate the general non-holonomic version of the RRT algorithm, which is able to determine an action that moves the agent towards a desired sample in state space via random sampling. We use the same version of this algorithm as described for example in~\cite{king2016}, which we recapitulate here in Alg.~\ref{alg:vanilla} and refer to as G-RRT.

\begin{algorithm}[t]
\caption{General-purpose non-holonomic RRT (G-RRT)}\label{alg:vanilla}
\begin{algorithmic}[1]
\Require Tree contains root node; $N \gets 1$
\While{$N<N_{max}$} \label{line:mainloop}
\State $\bm{x}_{sample} \gets$ random point in state space
\State $\bm{x}_{node} \gets$ node closest to $\bm{x}_{sample}$ currently in tree
\State $d_{min} \gets \infty; \bm{x}_{new} \gets $ NULL
\While{$k<K_{max}$} \label{line:loop}
\State $\bm{a} \gets$ random action
\State $\bm{x}_a \gets F(\bm{x}_{node},\bm{a})$
\If{Stable($\bm{x}_a$) \textbf{and} $dist(\bm{x}_{sample},\bm{x}_a) < d_{min}$ }\label{line:stable}
\State $d_{min} \gets dist(\bm{x}_{sample},\bm{x}_a)$
\State $\bm{x}_{new} \gets \bm{x}_a$
\EndIf
\State $k \gets k+1$
\EndWhile
\If{$\bm{x}_{new}$ is not NULL}
\State Add $\bm{x}_{new}$ to tree with $\bm{x}_{node}$ as parent
\State $N\gets N+1$
\EndIf
\EndWhile
\end{algorithmic}
\end{algorithm}

The essence of this algorithm is the \textbf{while} loop in line~\ref{line:loop}: it is able to grow the tree in a desired direction by sampling a number $K_{max}$ of random actions, then using the transition function $F()$ of our problem to evaluate which of these produces a new node that is as close as possible to a sampled target. 

Our only addition to the general-purpose algorithm is the stability check in line \ref{line:stable}: a new node gets added to the tree only if it passes a stability check. This check consists of advancing the simulation for an additional 1s with no change in the action; if, at the end of this interval, the object has not been dropped (i.e. the height of the object is above a threshold) the new node is deemed stable and added to the tree. Assuming a typical simulation step of 2 ms, this implies 500 additional calls to $F()$ for each sample; however, it does away with the need for domain-specific analytical stability methods as we used for M-RRT.

Overall, the great advantage of this algorithm lies in its simplicity and generality. The only manipulation-specific component is the aforementioned stability check. However, its performance can be dependent on $K_{max}$ (i.e. number of action samples at each iteration), and each of these samples requires a call to the transition function. This problem can be alleviated by the advent of highly efficient and massively parallel physics engines implementing the transition function, which is an important research direction complementary to our study. 

\subsection{Reinforcement Learning}
\label{sec:RL}

While the RRT algorithms we have discussed so far have excellent abilities to explore the complex state space of in-hand manipulation, and to identify (approximate) transitions that follow the complex manifold structure of this space, they do not provide directly usable policies. In fact, M-RRT does not provide actions to use, and the transitions might not be feasible under the true transition function. G-RRT does find transitions that are valid, and also identifies the associated actions, but provides no mechanism to act in states that are not part of the tree, or to act under slightly different transition functions.

In order to generate closed-loop policies able to handle variability in the encountered states, we turn to RL algorithms. Critically, we rely on the trees generated by our sampling-based algorithms to ensure effective exploration of the state space during policy training. The specific mechanism we use to transition information from the sampling based tree to the policy training method is via the reset distribution: we select relevant paths from the planned tree, then use the nodes therein as reset states for policy training.

We note that the sampling-based trees as described here are task-agnostic. Their effectiveness lies in achieving good coverage of the state space (usually within pre-specified limits along each dimension). Once a specific task is prescribed (e.g. via a reward function), we must select the paths through the tree that are relevant for the task. For the concrete problem chosen in this paper (large in-hand object reorientation) we rely on the heuristic of selecting the top ten paths from the RRT tree that achieve the largest angular change for the object around the chosen rotation axis. (Other selection mechanisms are also possible; a promising and more general direction for future studies is to select tree branches that accumulate the highest reward.) After selecting the task-relevant set of states from the RRT tree, we use a uniform distribution over these states as a reset distribution for RL. 

Our approach is compatible with RL methods that alternate between collecting episode rollouts and updating the policy, and restarting episode rollouts starting from a new set of states. Thus, both off-policy and on-policy RL are equally feasible. However, we use on-policy learning due to its its compatibility with GPU physics simulators and relative training stability.  

\section{Experiments \& Results}
\label{sec:results}

\subsection{Experimental Setup}

We use the robot hand shown in Fig.~\ref{fig:eyecandy}, consisting of five identical fingers. Each finger comprises a roll joint and two flexion joints, for a total of 15 fully actuated position-controlled joints. For the real hardware setup, each joint is powered by a Dynamixel XM430-210T servo motor. The distal link of each finger consists of an optics-based tactile fingertip as introduced by \textcite{Piacenza2020-zp}. 

\begin{figure}
    \centering
    \includegraphics[trim=0mm 20mm 0mm 0mm,clip,width=0.3\textwidth]{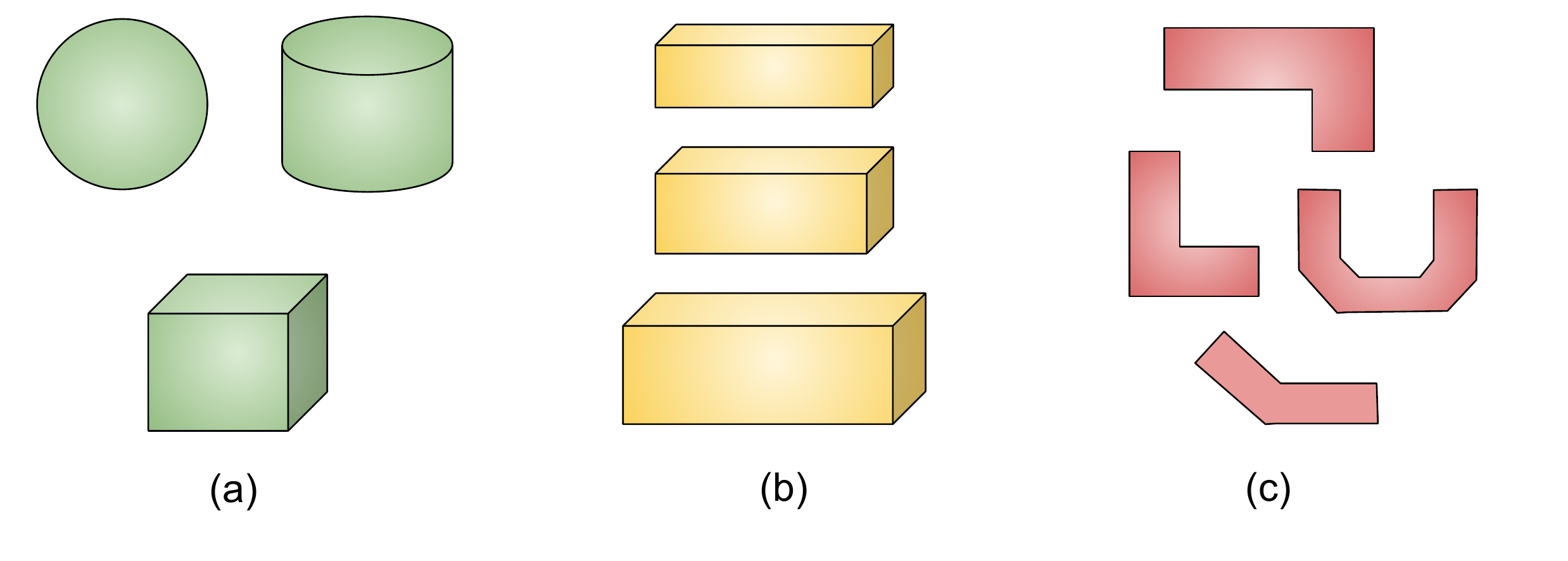}
    \caption{The object shapes for which we learn finger-gaiting. From left to right: the easy, moderate and hard categories.}
    \label{fig:objset}
\end{figure}

We test our methods on the object shapes illustrated in Fig.~\ref{fig:objset}. We split this into categories: "easy" objects (sphere, cube cylinder), "moderate" objects (cuboids with elongated aspect ratios), and "hard" objects (either concave L- or U-shapes). 
We note that in-hand manipulation of the objects in the "hard" category has not been previously demonstrated in the literature.

\subsubsection{Exploration Trees Setup}

\begin{figure}[t!]
    \centering
    \includegraphics[width=0.45\textwidth]{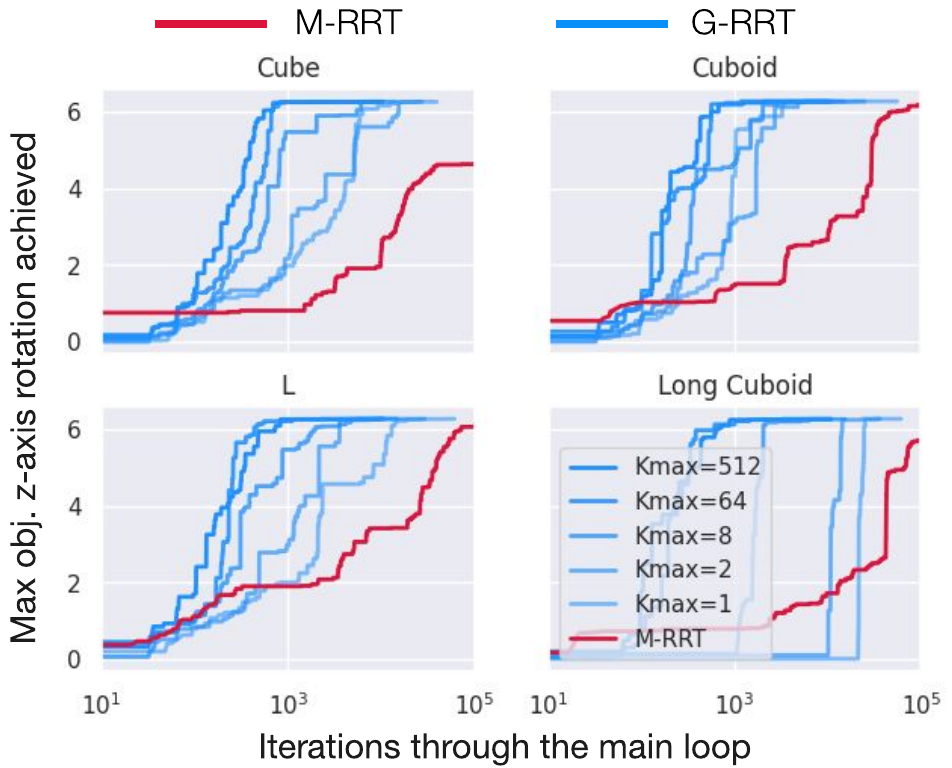}
    \caption{Tree expansion performance for G-RRT and M-RRT. We plot the number of attempted tree expansions (i.e. iterations through the main loop, on a log scale) against the maximum object z-axis rotation achieved by any tree node so far. For G-RRT, we also plot performance for different values of $K_{max}$, the number of random actions tested at each iteration.}
    \label{fig:rrtcomp}
\end{figure}

We run both G-RRT and M-RRT on the objects in our set. The first test consists, for both algorithms, in how effectively the tree explores its available state space given the number of iterations through the main loop (i.e. the number of attempted tree expansions towards a random sample). As a measure of tree growth, we look at the maximum object rotation achieved around our target axis. (We note that any rotation beyond approximately $\pi/4$ radians can not be done in-grasp, and requires finger repositioning.) Thus, for both algorithms, we compare maximum achieved object rotation vs. number of expansions attempted (on log scale). The results are shown in Fig.~\ref{fig:rrtcomp}. 

We found that both algorithms are able to effectively explore the state space. G-RRT is able to explore farther with fewer iterations, and its performance further increases with the number of actions tested at each iteration. We attribute this difference to the fact that M-RRT is constrained to taking small steps due to the linearization of constraints used in the extension projection. G-RRT, which uses the actual physics of the domain to expand the tree, is able to take larger steps at each iteration without the risk of violating the manipulation constraints.

As expected, the performance of G-RRT improves with the number $K_{max}$ of actions tested at each iteration. Interestingly, the algorithm performs well even with $K_{max}=1$; this is equivalent to a tree node growing in a completely random direction, without any bias towards the intended sample. However, we note that, at each iteration, the node that grows is the closest to the state-space sample taken at the beginning of the loop. This encourages growth at the periphery of the three and along the real constraint manifolds, and, as shown here, still leads to effective exploration.

Both these algorithms can be parallelized at the level of the main loop (line \ref{line:mainloop}). However, the extensive sampling of possible actions, which is the main computational expense of G-RRT (line \ref{line:loop}) also lends itself to parallelization. In practice, we use the IsaacGym \cite{Makoviychuk2021-ko} parallel simulator to parallelize this algorithm at both these levels (32 parallel evaluations of the main loop, and 512 parallel evaluations of the action sampling loop). This made both algorithms practical for testing in the context of RL. \footnote{Given the advent of increasingly more powerful parallel architectures for general physics simulation, we expect that more general methods that are easier to parallelize might win out in the long term over more problem-specific solutions that are more sample efficient at the individual thread level.}

We then moved on to using paths from the planned trees in conjunction with RL training. Since our goal is finger gaiting for z-axis rotation, we planned additional trees with each method where object rotation around the x- and y-axes was restricted to 0.2 radians. Then, from each tree, we select $2 \times 10^4$ nodes from the paths which exhibit the most rotation around the z-axis, and extract their nodes. On average, each such path comprises 100-400 nodes. In the case of G-RRT, we recall that all tree nodes are subjected to an explicit stability check under full system dynamics before being added to the tree; we can thus use each of them as is. If using M-RRT, we also apply the same stability check to the nodes of the longest paths at this time, before using them as reset states for RL as described next.

\subsubsection{Reinforcement Learning Setup} We train our policies using Asymmetric Actor Critic PPO \cite{Pinto2017-nw, Schulman2017-fz}; all training is done in the IsaacGym simulator. The critic uses object pose $\boldsymbol{p}$, object velocity $\dot{\boldsymbol{p}}$, and net contact force on each fingertip $\boldsymbol{t}_1 \ldots \boldsymbol{t}_{m}$ as feedback in addition to the feedback already as input to the policy network. Similar to \textcite{Khandate2022-qt}, we use a reward function that rewards object angular velocity about z-axis if the hand re-orienting the object with at least three fingertip contacts. In addition, we include penalties for the object's translational velocity and its deviation from the initial position \cite{Qi2022-wy}. We also use early termination to terminate the episode rollout if there are fewer than two contacts. 

\subsection{Experimental Conditions and Baselines}
\begin{figure*}
    \centering
    \includegraphics[width=\textwidth]{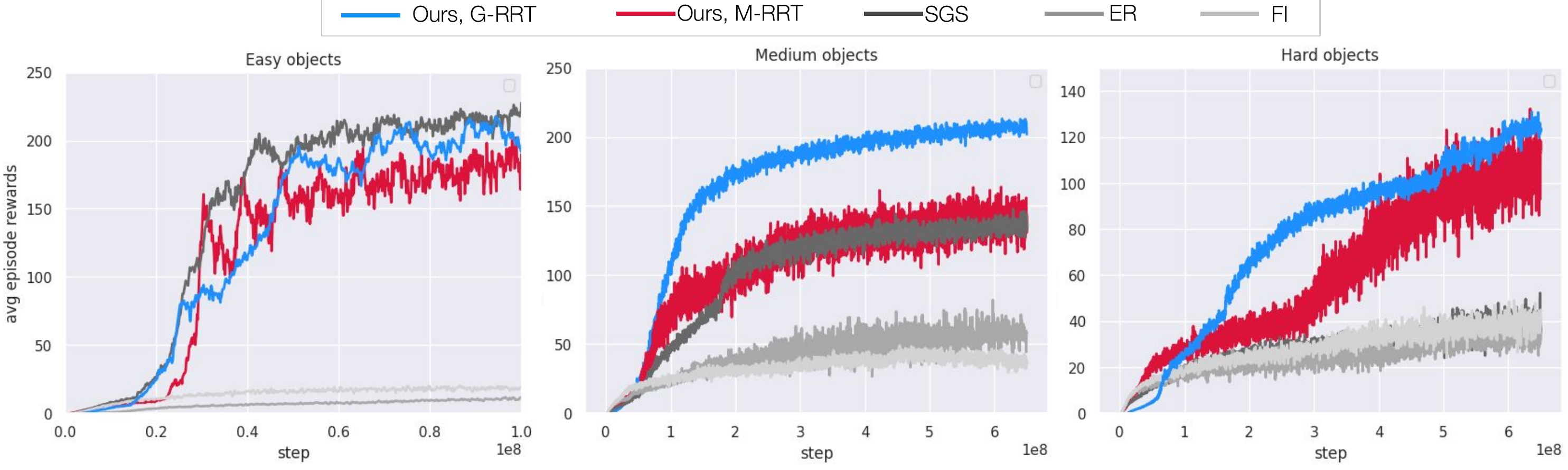}
    \caption{Training performance for our methods (G-RRT and M-RRT) and a number of baselines on the object categories shown in Fig.~\ref{fig:objset}.}
    \label{fig:results}
\end{figure*}
In our experiments, we compare the following approaches:

\subsubsection{Ours, G-RRT} In this variant, we use the method presented in this paper, relying on exploratory reset states obtained by growing the tree via G-RRT. In all cases, we use a tree comprising $10^5$ nodes as informed the ablation study in Fig~\ref{fig:treesize}.

\subsubsection{Ours, M-RRT} This is also the method presented here, but using M-RRT for exploration trees. Again, we use trees comprising $10^5$  nodes. 

\subsubsection{Stable Grasp Sampler (SGS)} This baseline represents an alternative to the method presented in this paper: we use a reset distribution consisting of stable grasps generated by sampling random joint angles and varying object orientation about the rotation axis. This approach has been demonstrated precision in-hand manipulation with only intrinsic sensing \cite{Khandate2022-qt, Qi2022-wy} for simple shapes. 

\subsubsection{Explored Restarts (ER)} This method selects states explored by the policy itself during random exploration to use as reset states~\cite{Tavakoli2018-ah}. It is highly general, with no manipulation-specific component, and requires no additional step on top of RL training. We implement the "uniform restart" scheme as it was shown to have superior performance on high dimensional continuous control tasks. However, we have found it to be insufficient for the complex state space of our problem: it fails to learn a viable policy even for simple objects.

\subsubsection{Fixed Initialization (FI)} For completeness, we also tried restarting training from a single fixed state. As expected, this method also failed to learn even in the simple cases. Additionally, we evaluated fixed initialization with gravity curriculum (zero to full). The policy only learned in-grasp manipulation, reorienting the object by the maximum possible amount without breaking contact before dropping. We found that the policy did not learn finger-gaiting even with zero gravity when using a fixed initialization. Thus, fixed initialization with or without gravity curriculum learning does not help with learning finger-gaiting. We hypothesize that curriculum learning has limited power to address exploration issues because policies tend to converge to sub-optimal behaviors that are hard to overcome later in training.

\subsection{Results}

Training results are summarized Fig.~\ref{fig:results}. The performance on easy objects confirms the results of previous studies, which showed that a reset distribution consisting of random grasps (SGS) enables learning of rotation gaits; sampling-based exploration (our methods) achieves similar performance. For medium objects, G-RRT, M-RRT, and SGS again all learn to gait, but the policies learned via G-RRT exploration are more effective. Finally, for complex problems (hard objects), a random grasp-based reset distribution is no longer workable. Only G-RRT and M-RRT are able to learn manipulation, and G-RRT does so more efficiently. We also note that none of the domain-agnostic methods (ER and FI) are able to learn in-hand manipulation on any object set, in the allotted training time. 

\begin{figure}[t!]
    \centering
    \includegraphics[width=0.45\textwidth]{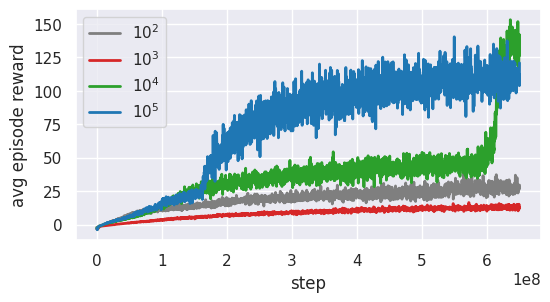}
    \caption{Training performance with different tree sizes. The training curves shown are for different set of reset states obtained from trees of varying sizes. We see that we need a sufficiently large tree with at least $10^4$ nodes to enable learning. However, training is most reliable with $10^5$ nodes.}
    \label{fig:treesize}
\end{figure}

We also studied the impact of size of the tree used in extracting reset states. Fig~\ref{fig:treesize} summarizes our results for learning a policy for an L-shaped object using tree of different sizes grown via G-RRT. Qualitatively, we observe that, as the tree grows larger, the top 100-400 paths sampled from the tree contain increasingly more effective gaits, likely closer to the optimal policy. This suggests a strong correlation between the optimality of states used for reset distribution and sample efficiency of learning.

\begin{figure}[t!]
    \centering
    \includegraphics[width=0.45\textwidth]{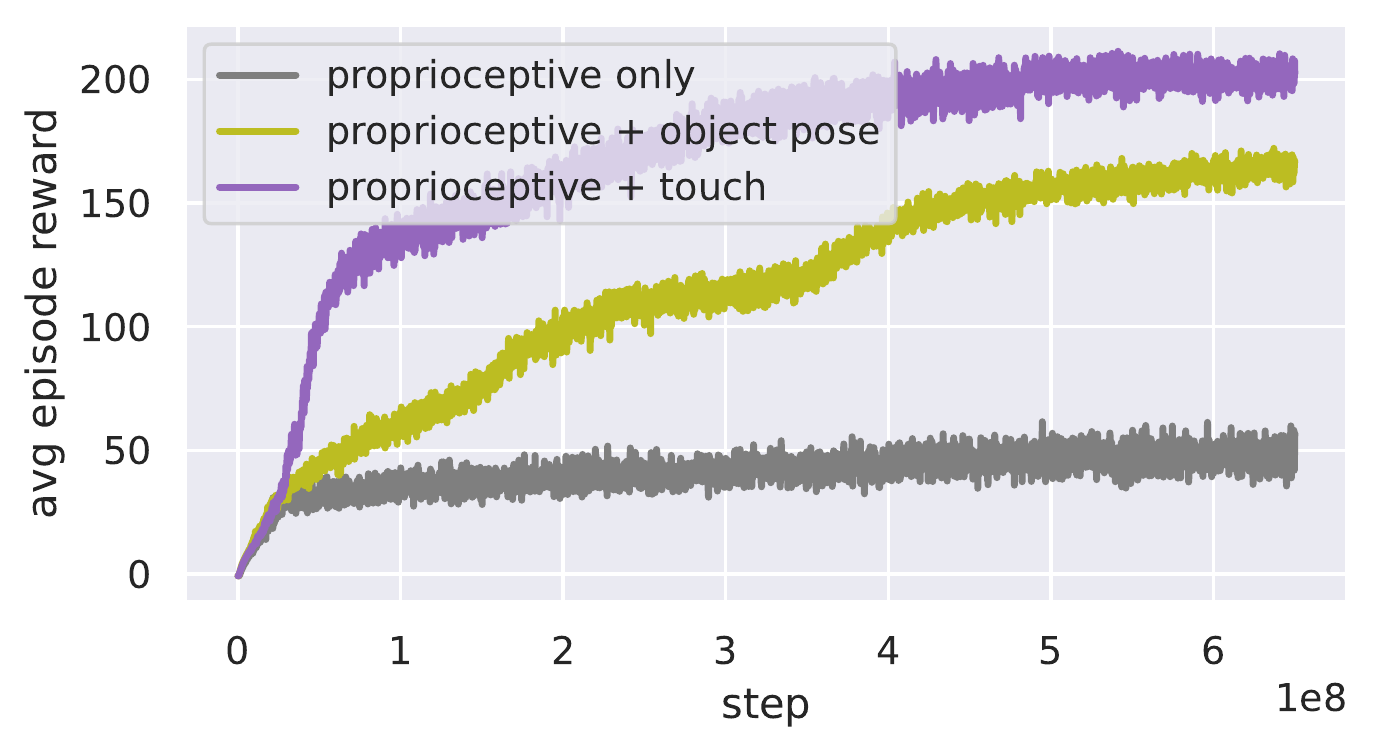}
    \caption{Ablation of policy feedback components for L-shaped object. We note that touch feedback is essential in the absence of object pose feedback, and also leads to faster learning in comparison with object pose feedback.}
    \label{fig:sense_abltn}
\end{figure}

In addition, we performed an ablation study of policy feedback. Particularly, we aimed to compare intrinsically available tactile feedback vs. object pose feedback that would require external sensing. Fig.~\ref{fig:sense_abltn} summarizes these results for an L-shaped object. First, we found that touch feedback is essential for all moderate and hard objects in the absence of object pose feedback. For these objects, we also saw that replacing this tactile feedback with object pose feedback results in slower learning, underscoring the importance of touch feedback for in-hand manipulation skills. Richer tactile feedback such as contact position, normals, and force magnitude can be expected to provide even stronger improvements; we hope to explore this in future work.

\subsection{Evaluation on real hand}

To test the applicability of our method on real hardware, we attempted to transfer the learned policy for a subset of representative objects: cylinder, cube, cuboid \& L-shape. We chose these objects to span the range from simpler to more difficult manipulation skills.

For sim-to-real transfer, we take a number of additional steps. We impose velocity and torque limits in the simulation, mirroring those used on the real motor ($0.6$ rad/s and $0.5$ N-m respectively). We found that our hardware has significant latency of $0.05$s which we included in the simulation. In addition, we modified the angular velocity reward to maintain a desired velocity instead of maximizing the object's angular velocity. We also randomize joint origins ($0.1rad$), friction coefficient ($1-40$), and train with perturbation forces ($1N$). All these changes are introduced successively via a curriculum.  

For sensing, we used the current position and setpoint from the motor controllers with no additional changes. For tactile data, we found that information from our tactile fingers is most reliable for contact forces above 1 N. We thus did not use reported contact data below this threshold, and imposed a similar cutoff in simulation. Overall, we believe that a key advantage of exclusively using proprioceptive data is a smaller sim-to-real gap compared to extrinsic sensors such as cameras. 

\begin{figure*}
    \centering
    \includegraphics[width=0.8\textwidth]{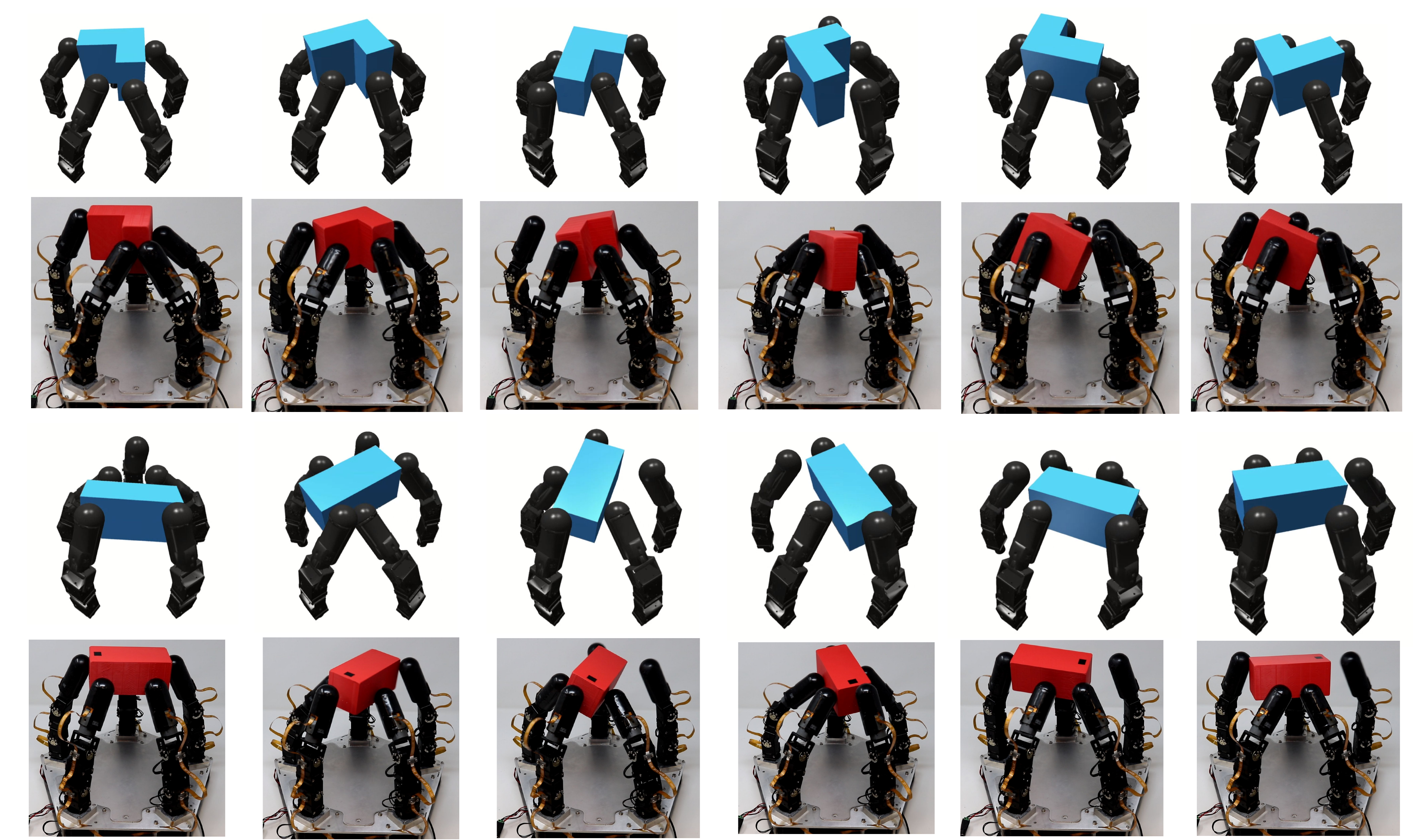}
    \caption{Key frames of the finger-gaiting in simulation and on the real hand for representative objects in simulation and on real hand. Representative videos of these tasks can be found on our project website \href{sbrl.cs.columbia.edu}{sbrl.cs.columbia.edu}}
    \label{fig:gaiting}
\end{figure*}

For the set of representative objects, we ran the respective policy ten consecutive times, and counted the number of successful complete object revolutions achieved before a drop. In other words, five revolutions means the policy successfully rotated the object for $1,800^{\circ}$ before dropping it. 
In addition, we also report the average object rotation speed observed during the trials. The results of these trials are summarized in Table~\ref{tab:sim2real}. Fig.~\ref{fig:gaiting} shows the keyframes of the finger-gaiting we achieved by the policy on the hand and also compared it with manipulation observed in simulation.

\section{Discussion and Conclusions}

The results we have presented show that sampling-based exploration methods make it possible to achieve difficult manipulation tasks via RL. In fact, these popular and widely used classes of algorithms are highly complementary in this case. RL is effective at learning closed-loop control policies that maintain the local stability needed for manipulation, and, thanks to training on large number of examples, are robust to variations in the encountered states. However, the standard RL exploration techniques (random perturbations in action space) are ineffective in the highly constrained state space with complex manifold structure of manipulation tasks. Conversely, SBP methods, which rely on a fundamentally different approach to exploration, can effectively discover relevant regions of the state space, and convey this information to RL training algorithms, for example via an informed reset distribution.

\begin{table}
    \centering
    \caption{Manipulation performance in simulation vs. real hardware. We report median number of object rotations achieved before dropping the object in ten consecutive trials, as well as the time needed to perform these rotations.}
    \ra{1.3}
    \begin{tabular}{ccc}
        \midrule
         \phantom{} & Median revolutions &  Mean rotation speed (rad/s)\\
         \midrule
         Cylinder & 5 &  0.42 \\
         Cube (s) &  4.5 & 0.44 \\
         Cuboid & 1.5  & 0.44 \\
         L-shape &  1.5 & 0.24 \\
         \midrule
    \end{tabular}

    
    \label{tab:sim2real}
\end{table}

Since sampling-based exploration methods are not expected to generate directly usable trajectories, exploration can also use approximate models of physical constraints, which can be informed by well-established analytical models of robotic manipulators. Interestingly, we found that using the general-purpose exploration algorithm using the full transition function of the environment is still more sample-efficient than using such analytical constraint models. Nevertheless, both are usable in practice, particularly with the advent of massively parallel physics simulators.

We use this approach to demonstrate finger gaiting precision manipulation of both convex and non-convex objects, using only tactile and proprioceptive sensing. Using only these types of intrinsic sensors makes manipulation skills insensitive to occlusion, illumination or distractors, and reduces the sim-to-real gap. We take advantage of this by demonstrating our approach both in simulation and on real hardware. We note that, while some applications naturally preclude the use of vision (e.g. extracting an object from a bag), we expect that in many real-life situations future robotic manipulators will achieve the best performance by combining touch, proprioception and vision.

Learning in-hand object reorientation to achieve a given desired pose may also potentially benefit from our approach, for example by leveraging information from the RRT tree to find appropriate trajectories for reaching a specific node. Another more general and promising direction for future work involves other mechanisms by which ideas from sampling-based exploration can facilitate RL, beyond reset distributions. Some SBP algorithms can also be used to suggest possible actions for transitions between regions of the state space, a feature that we do not take advantage of here (even though one of the exploration algorithms we use does indeed compute actions). Alternatively, sampling-based exploration  techniques could be integrated directly into the policy training mechanisms, removing the need for two separate stages during training. We hope to explore all these ideas in future work.
\\~\\
\noindent \textit{Acknowledgements.} This work was supported in part by the Office of Naval Research grant N00014-21-1-4010 and the National Science Foundation grant CMMI-2037101.


\newpage
\printbibliography
\end{document}